\documentclass[table]{article}

\usepackage[preprint]{neurips_2025}

\usepackage[utf8]{inputenc} % allow utf-8 input
\usepackage[T1]{fontenc}    % use 8-bit T1 fonts
\usepackage{hyperref}       % hyperlinks
\usepackage{url}            % simple URL typesetting
\usepackage{booktabs}       % professional-quality tables
\usepackage{amsfonts}       % blackboard math symbols
\usepackage{nicefrac}       % compact symbols for 1/2, etc.
\usepackage{microtype}      % microtypography
\usepackage{xcolor}         % colors
\usepackage{pifont}
\usepackage{makecell}
\usepackage{multirow} 
\usepackage{booktabs}
\usepackage{multirow}
\usepackage{tabularx}
\usepackage{graphicx}
\usepackage{subfigure}
\usepackage[most,skins,theorems]{tcolorbox}
\usepackage{cleveref}
\usepackage{wrapfig}

\crefname{figure}{Fig.}{Fig.}
\crefname{table}{Tab.}{Tab.}
\crefname{section}{Sec.}{Sec.}
\crefname{appendix}{App.}{App.}

\tcbset{
  aibox/.style={
    width=\linewidth,
    top=8pt,
    bottom=4pt,
    colback=blue!6!white,
    colframe=black,
    colbacktitle=black,
    enhanced,
    center,
    attach boxed title to top left={yshift=-0.1in,xshift=0.15in},
    boxed title style={boxrule=0pt,colframe=white,},
  }
}

\newtcolorbox{AIbox}[2][]{aibox,title=#2,#1}

\title{\textsc{Knowledge} \includegraphics[height=1em]{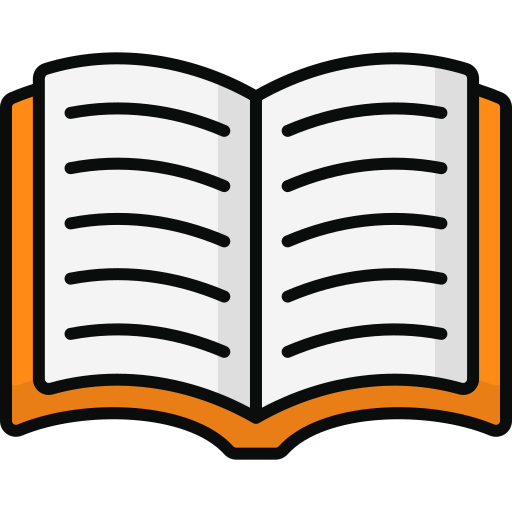} or \textsc{Reasoning} \includegraphics[height=1em]{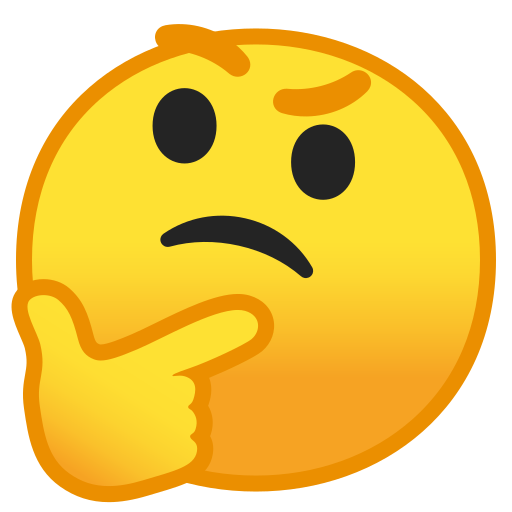} ? \\ A Close Look at How LLMs Think Across Domains}

\author{%
  Juncheng Wu$^{1\star}$ \quad
  Sheng Liu$^{2\star}$ \quad
  Haoqin Tu$^{1\star}$ \quad
  Hang Yu$^{3\star}$ \quad
  Xiaoke Huang$^{1}$ \quad \\ \vspace{.1em}
  \textbf{James Zou}$^{2}$ \quad
  \textbf{Cihang Xie}$^{1}$ \quad
  \textbf{Yuyin Zhou}$^{1}$ \vspace{.3em}
  \\\small $^{\star}$equal technical contribution\vspace{.5em} \\
  $^{1}$UC Santa Cruz \, \quad
  $^{2}$Stanford University \, \quad
  $^{3}$Tongji University \vspace{.5em}
  \\
  \small
  \hspace{3em} \includegraphics[height=1.1em]{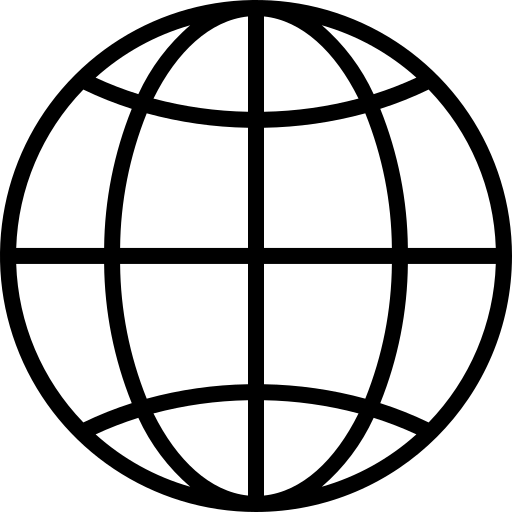} \textbf{Project Page}: \url{https://ucsc-vlaa.github.io/ReasoningEval} \\
  \small
  \hspace{3em} \includegraphics[height=1.2em]{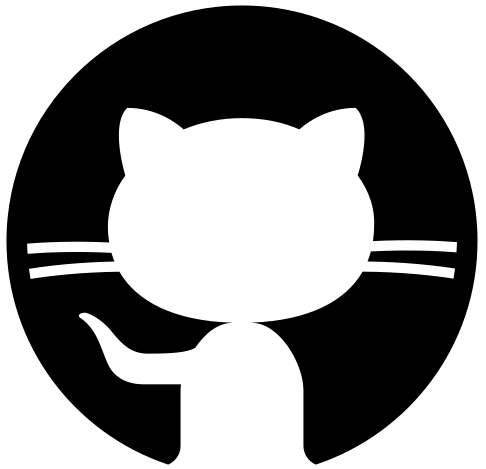} \textbf{Code}: \url{https://github.com/UCSC-VLAA/ReasoningEval}  \\
}

\begin{document}

\maketitle

\begin{abstract}

Recent advances in reasoning-enhanced Large Language Models such as OpenAI-o1/3 and DeepSeek-R1 have significantly improved performance on complex tasks. However, the quality and transparency of their internal reasoning processes remain underexplored. 
This work moves beyond the final-answer accuracy and investigates step-by-step reasoning in the medical and mathematical domains by explicitly decomposing the thinking trajectories into two parts: \emph{knowledge} and \emph{reasoning}. Specifically, we introduce a fine-grained evaluation framework that judges: (1) the correctness of \emph{knowledge} used (measured by Knowledge Index (KI)) and (2) the quality of \emph{reasoning} (measured by Information Gain (InfoGain)).
Using this framework, we study R1-distilled and base Qwen models trained with 
% We conduct a case study on R1-distilled and the base Qwen models with 
supervised fine-tuning (SFT) and/or reinforcement learning (RL) in the medical and math domains. Three intriguing findings emerge: 
(1) The general reasoning abilities in R1-distilled models do not transfer effectively to the medical domain through either SFT or RL.
(2) SFT raises final-answer accuracy in both domains, but often at the cost of reasoning quality: InfoGain drops by 38.9\% on average compared with untrained models; In the medical domain, however, SFT remains crucial because domain knowledge is indispensable.
% while SFT improves final accuracy in both domains, it often compromises reasoning, as reflected by average 38.9\% lower InfoGain scores against untrained models---yet it remains essential in the medical domain, where domain knowledge is critical for accuracy; and
(3) RL enhances medical reasoning by pruning inaccurate or irrelevant knowledge from reasoning paths, thereby improving both reasoning accuracy and knowledge correctness.
% We hope this work can encourage future research toward more reliable LLM reasoning.
\end{abstract}
\section{Introduction}
\label{sec:intro}
Recent proprietary and open-source Large Language Models (LLMs)~\citep{deepseek-aiDeepSeekR1IncentivizingReasoning2025,ji2025amthinking,jurayj2025finalanswer} have demonstrated remarkable progress in reasoning-intensive benchmarks, particularly in mathematics and general knowledge~\citep{yangQwen3TechnicalReport2025,seedSeed15ThinkingAdvancingSuperb2025,kimiteamKimiK15Scaling2025,bercovichLlamaNemotronEfficientReasoning2025}. Despite this rapid progress, the evaluation of LLM reasoning remains limited in scope—largely centered on final-answer accuracy or aggregate performance metrics. Such evaluations obscure the step-by-step process by which models reason and offer little insight into the interplay between factual knowledge and logical inference that underlies these capabilities.

Earlier work~\cite{golovneva2022roscoe} evaluates reasoning based on its embedding similarity to the original question, assuming higher similarity implies greater informativeness and faithfulness. 
However, LLMs often rely on internal knowledge or previous deductions, making question alignment an unreliable measure of knowledge accuracy or reasoning quality.
As shown in~\cref{tab:ROSCOE_res} in the Appendix, existing reasoning metrics yield similar scores across models but with differing capacities, suggesting their unreliability.

\begin{figure}[t!]
    \centering
    \includegraphics[width=\linewidth]{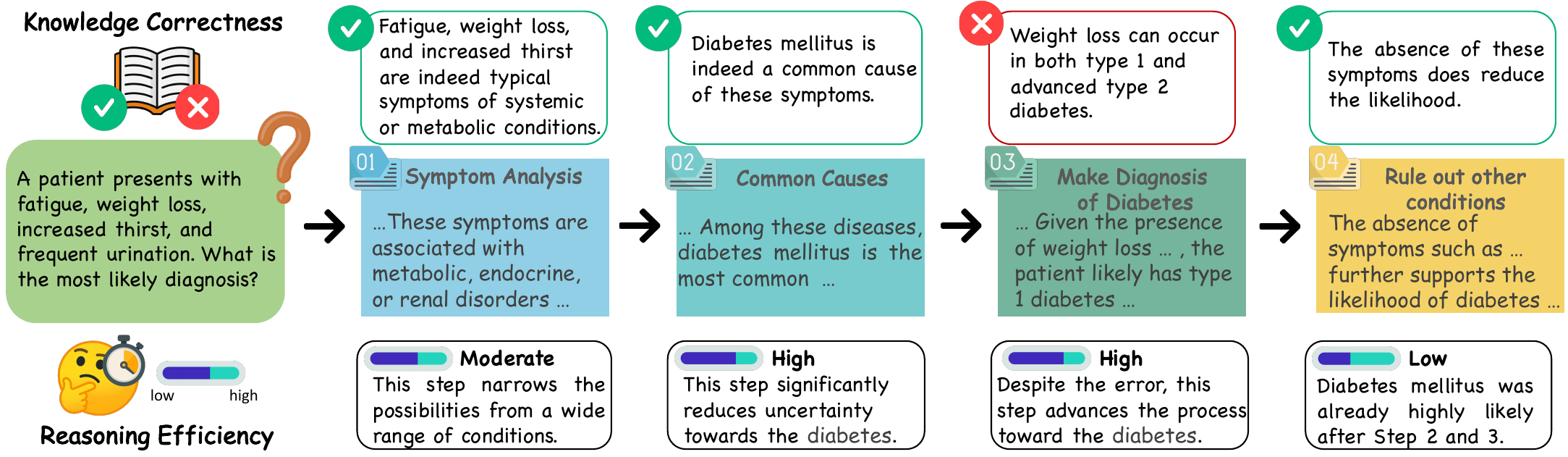}
    \caption{\textbf{Reasoning and Knowledge are Different Evaluation Aspects for LLMs.} A reasoning step may effectively reduce uncertainty toward the final answer despite relying on incorrect knowledge (\emph{e.g.}, Step 3), or it may present factually correct but irrelevant/redundant knowledge that hinders reasoning efficiency (\textit{e.g.}, Step 4). Accuracy alone fails to capture these nuances. We introduce two complementary metrics that separately evaluate knowledge correctness and reasoning informativeness.}
    \vspace{-.5em}
    \label{fig:teaser}
\end{figure}

Such gaps in reasoning evaluation are particularly salient when comparing domains with different knowledge and reasoning demands. 
For instance, mathematical problems often emphasize symbolic manipulation and internal consistency~\cite{ahn2024large}, whereas medical tasks typically require the integration of domain-specific knowledge grounded in external facts~\cite{goh2024large}. 
Both domains involve multi-step reasoning, but they differ in how much they depend on knowledge versus the reasoning steps required during generation. 
Understanding these differences is critical not only for building domain-adaptive models but also for advancing interpretability and reliability in high-stakes applications.

\begin{figure*}[t!]
\centering
\includegraphics[width=0.92\linewidth]{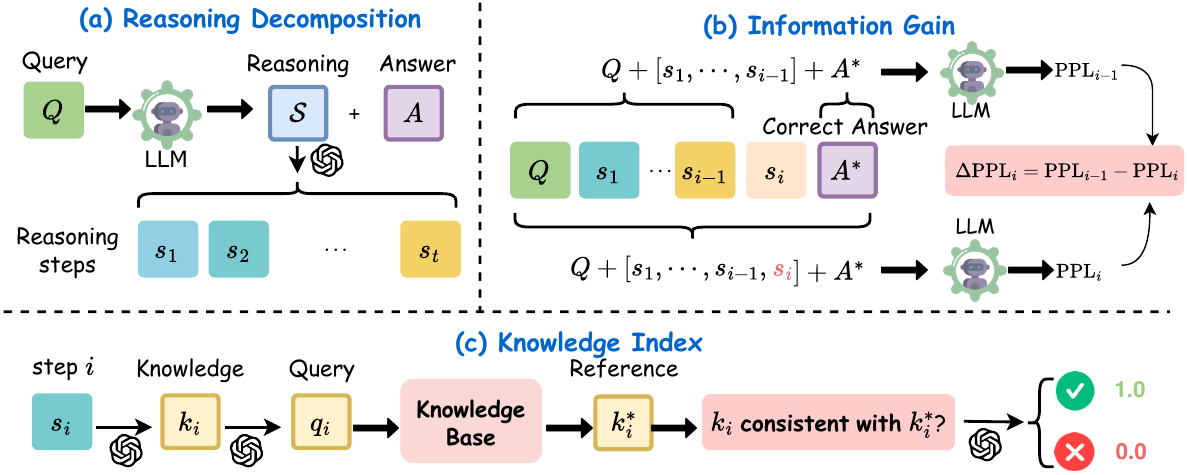} 
\caption{\textbf{Evaluation Pipeline.} (a) We decompose the model's reasoning into reasoning steps using \texttt{gpt4o}\cite{openai2024gpt4o}
, then evaluate the (b) information gain and (c) knowledge index of each reasoning step.}
\label{fig:pipeline}
\end{figure*}

In this work, we pose a fundamental question: \textit{What are the respective roles of knowledge and reasoning in the thinking process of LLMs, and how do they interact across different domains?} 
To answer this, we introduce an evaluation framework (see~\cref{fig:pipeline}) that decomposes each reasoning step into two components: the factual knowledge it invokes and the logical reasoning operation it performs. We define two novel metrics to quantify reasoning and knowledge:
(1) \textbf{Information Gain (Info Gain)} how much a reasoning step reduces uncertainty toward the final answer, calculated as the probability gap between adjacent response steps. 
A higher Info Gain indicates a more informative reasoning path towards the final answer.
(2) \textbf{Knowledge Index (KI)}, on the contrary, evaluates the factual correctness of each step by verifying extracted knowledge against external ground truth sources.
We identify the knowledge point in each reasoning step and access external factual data to verify if the knowledge aligns with the retrieved facts. 
Finally, models with stronger knowledge grounding yield higher KI scores. 
A running example is provided in Figure~\ref{fig:teaser} to explain our motivation intuitively.
This fine-grained evaluation allows us to characterize not just the model's final performance, but also the trajectory it takes to get there.

Building upon this framework, we analyze models trained via supervised fine-tuning (SFT) and reinforcement learning (RL) across both mathematical and medical domains.
Our findings reveal several key insights: (1) mathematical reasoning does not naturally transfer to the medical domain via SFT, largely due to domain-specific knowledge gaps, as evidenced by the consistently lower performance of the DeepSeek-distilled model; (2) tasks across domains demand distinct model competencies---medical problems require richer domain knowledge, with knowledge–accuracy correlations exceeding those of reasoning–accuracy in four of five benchmarks; (3) while SFT improves final accuracy and raises knowledge levels (\textit{e.g.}, a 6.2\% average KI increase on medical tasks), it often introduces verbose or suboptimal reasoning, reducing Info. Gain by an average of 38.9\%; and (4) RL mitigates such inefficiencies by reinforcing correct knowledge trajectories, boosting medical knowledge with an average KI gain of 12.4.
We hope our study can foster a more comprehensive understanding of individual LLM reasoning steps and how they collectively influence the reliability of model outcomes. 
This, in turn, could be helpful in guiding the way towards more effective training strategies to build reliable LLMs.

\section{Related Works}
\paragraph{Reasoning-enhanced LLMs.}
Recent reasoning large language models (LLMs) have shown remarkable performance in the fields of mathematics~\cite{deepseek_r1,kimiteamKimiK15Scaling2025} and medicine~\cite{xie2024preliminary,chen2024huatuogpt}
Subsequent studies have focused on enhancing the reasoning abilities of LLMs by producing high-quality training datasets~\cite{wu2025medreason,huang2025m1,wang2025star} or crafting comprehensive reward mechanisms~\cite{chen2025sft,aggarwal2025l1}. 
However, these efforts are directed exclusively towards boosting the answering accuracy of reasoning LLMs but overlook the comprehension of internal reasoning processes.
This study seeks to investigate deeper into the reasoning process by introducing a step-by-step pipeline for reasoning evaluation, aiming to offer empirical findings that aid in the advancement of stronger reasoning models.

\paragraph{Evaluating LLMs Reasoning beyond Accuracy.}
Examining the quality of LLMs' reasoning cannot be restricted to just evaluating the accuracy of the final answer.
Prior studies concentrated on identifying factual inaccuracies throughout the whole reasoning process~\cite{zhao2023verify,xue2023rcot,prasad2023receval}, rather than assessing each reasoning step separately.
Research such as \cite{golovneva2022roscoe} proposed to evaluate an individual reasoning step by measuring its embedding similarity with the source information provided in the question.
However, similarity to the source content cannot fully reveal the knowledge correctness or logical effectiveness of a reasoning step, as LLMs often generate reasoning based on its internal knowledge and or previous reasoning steps.
Nonetheless, LLMs frequently produce reasoning grounded in their internal knowledge or prior reasoning steps. Merely resembling the original content doesn't entirely indicate the knowledge correctness or logical soundness of a reasoning step.
In this paper, we propose to step-by-step evaluate LLMs' across two dimensions to gain a thorough comprehension of the quality of reasoning.
\section{A Closer Look at Reasoning Evaluation}
\label{sec:method}

\subsection{Evaluation Pipeline}
All experiments in both the medical and math domains are initialized from the universal 7B-parameter base models \textbf{Qwen2.5‐7B}~\citep{qwen25_base} and \textbf{DeepSeek‐R1‐Distill‐Qwen 7B}~\citep{deepseek_r1}. 
We choose these two models as baselines because: (1) Qwen2.5-7B and its DeepSeek-distilled variant show strong generalization across domains, with the DeepSeek-distilled one often matching or outperforming larger or private models like Claude-3.5 in math and coding~\citep{deepseek_r1}; (2) their open-source nature enables in-depth study of training, evaluation, and architecture; and (3) their shared backbone ensures fair comparison of post-training thinking patterns.

We employ both the SFT and RL training to the model. In detail, for medical domain, we take \emph{huatuoGPT-o1}~\citep{chen2024huatuogpt} with SFT and RL data splits for respective SFT and RL training, while in the math domain, we employ SFT and RL splits from \emph{RLHFlow}~\citep{dong2024rlhfworkflow} for the corresponding training strategy.

With regard to the evaluation across both domains, we consider \href{https://medmcqa.github.io/}{MedMCQA}~\citep{pal2022medmcqa}, \href{https://paperswithcode.com/dataset/medqa-usmle}{MedQA‑USMLE}~\citep{jin2020medqa}, \href{https://pubmedqa.github.io/}{PubMedQA}~\citep{jin2019pubmedqa}, \href{https://huggingface.co/datasets/TIGER-Lab/MMLU-Pro}{MMLU‑Pro (Medical)}~\citep{wang2024mmlupro}, \href{https://arxiv.org/abs/2501.18362}{MedXpertQA}~\citep{zuo2025medxpert} for medical and \href{https://artofproblemsolving.com/wiki/index.php/AIME_Problems_and_Solutions}{AIME 2024}~\cite{aime2024}, \href{https://www.vals.ai/benchmarks/math500-03-24-2025}{MATH500}~\cite{vals2025math500}, \href{https://huggingface.co/datasets/hendrycks/competition_math}{AMC (10 \& 12)}~\cite{hendrycks2021math}, \href{https://github.com/EleutherAI/lm-evaluation-harness/tree/main/lm_eval/tasks/minerva_math}{Minerva‑Math}~\cite{lewkowycz2022minerva}, \href{https://www.maa.org/math-competitions}{USAMO 2025 / Olympiad}~\cite{usamo2025} for math. More details about model training configurations, evaluation and training datasets are in Appendix~\ref{app:training_details}.

In this section, we investigate the roles of reasoning and knowledge in model responses by first decomposing them into separate steps, and then introducing two novel metrics to evaluate the knowledge and reasoning abilities embedded in those responses.

\subsection{Response Decomposition}
In the first stage, we aim at decomposing the model responses into explicit and successive steps.
Given a question $Q$, the LLM produces the reasoning process $\mathcal{R}$ and subsequently delivers the answer $A$.
As depicted in \cref{fig:pipeline}~(a), our evaluation pipeline initially breaks down $\mathcal{R}$ into a sequence of logical steps using \texttt{gpt4o}, denoted as $\mathcal{S}=[s_1, s_2,\cdots, s_t]$. 
We employ different prompts fed in \texttt{gpt4o} for different domains with more details presented in \cref{fig:Medical_Reasoning_Decomposition} and \cref{fig:Math_Reasoning_Decomposition}.

\begin{figure*}[t]
\centering
\includegraphics[width=\linewidth]{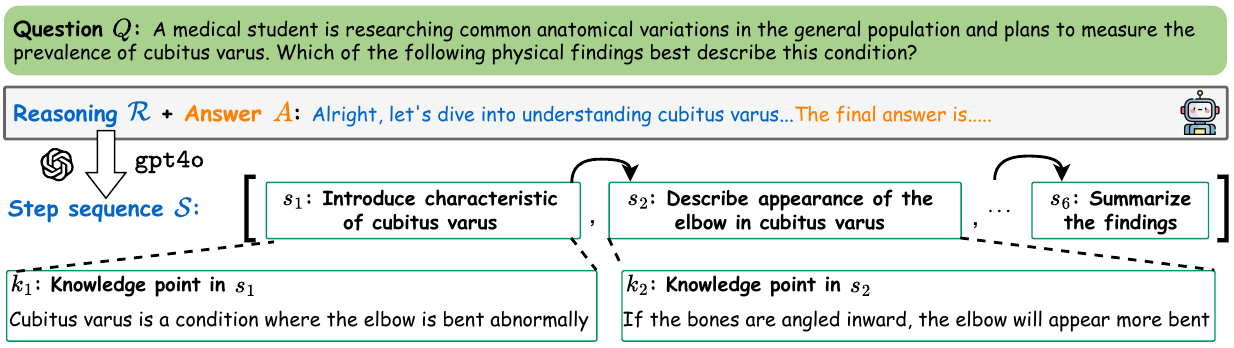} 
\caption{\textbf{Example of reasoning decomposition.} Every stage of reasoning corresponds to a logical step \(s_i\), accompanied by the specific knowledge point it contains (\(k_i\)). }
\label{fig:decompose_example}
\end{figure*}

For instance, as shown in \cref{fig:decompose_example}, to answer a question regarding cubitus varus, the model engages in reasoning involving six logical steps. The initial step introduces \textit{the characteristic of cubitus varus}, followed by \textit{describing the appearance of elbow} in the second step.

\subsection{Information Gain} \label{sec:info_gain}

To quantify the uncertainty reduced by each logical step towards the final answer, we track the change in the model’s confidence over each step, using the probability assigned to the true answer by leveraging step-wise perplexity (PPL)~\citep{brown2020language}. 
As shown in \cref{fig:pipeline}~(b), let $Q$ be the question, and $\mathcal{S} = [s_1, s_2, \dots, s_t]$ be the sequence of reasoning steps. Let $A^*$ be the correct solution to the query. After each step $s_i$, we concatenate the steps $s_{1:i}$ and compute the probability assigned to the correct solution. To avoid potential biases introduced by self-evaluation~\citep{panickssery2024llm}, we deploy another language model (\textit{i.e.,} the untrained Qwen2.5-7B) for PPL calculation:
\begin{align*}
    P_i = \Pi_{j=1}^N P(A_j^*|Q,s_{1:i}),
\end{align*}
where $A^*_j$ is the $j-$th token in the correct answer, and $N$ is the total number of tokens in $A^*$. 
This measures how likely the model is to generate the correct answer given the current plan. 
We then convert the probability into perplexity (PPL) and calculate the PPL gain between adjacent steps:
% \[
% [\text{PPL}_0, \text{PPL}_1, \dots, \text{PPL}_t],
% \] 
\[[\Delta\text{PPL}_1,\Delta\text{PPL}_2, \dots, \Delta\text{PPL}_t],\; \text{where } \Delta \text{PPL}_i = \text{PPL}_{i-1} - \text{PPL}_{i},\]
where $\text{PPL}_i$ represents the PPL of the $i$-th step. Each $\Delta\text{PPL}_i$ is designed to measure how effective a step $s_i$ is in reducing the model's uncertainty towards the final answer. 
In the end, we average the $\Delta\text{PPL}$s across all steps to obtain the final information gain:
\begin{align*}
    \Delta I = \frac{1}{t} \sum_{i=1}^t\Delta \text{PPL}_i.
\end{align*}
While a higher $\Delta I$ indicates that more information emerges during reasoning, reflecting stronger reasoning capabilities, a lower $\Delta I$ suggests weaker reasoning, likely characterized by redundant or less informative model responses.

\subsection{Knowledge Index} \label{sec:knowledge_index}
As for accessing the knowledge presented in model responses, we propose the metric Knowledge Index (KI) (show in \cref{fig:pipeline}~(c)). 
For each step, we retrieve ground truth answers for the extracted knowledge from medical textbooks~\cite{jin2020medqa} using \texttt{gpt-4o}, and then use these references to assess whether the step aligns with the factual knowledge.

Specifically, we decompose the process of obtaining the knowledge correctness into three stages: 
\begin{itemize}
    \item \textbf{Knowledge Extraction.} In each step $s_i$, we employ \texttt{gpt4o}~\citep{openai2024gpt4o} to extract knowledge in the model response step $s_i$, denoted as $k_i$. We then reformat $k_i$ as a question of querying about specific knowledge for the next step retrieval from database. 
    
    \item \textbf{Knowledge Retrieval.} Using the generated knowledge query, we employ a widely-employed and well-structured medical database to retrieve relevant external knowledge $k^*_i$ as the ground truth answer to the knowledge query $k_i$. 
    
    \item \textbf{Knowledge Judgement.} Finally, to determine whether the extracted knowledge aligns with the ground truth, we use \texttt{gpt-4o} to assess whether each reasoning step $s_i$ is consistent with the retrieved facts provided in the prompt. 
    The outcome of this consistency check, $\text{consistency}_i$, is recorded as a Boolean value: \texttt{True} if consistent, \texttt{False} otherwise.
\end{itemize}
The overall knowledge index across all steps for each query is computed as 
\begin{align*}
    \text{KI} = \frac{1}{t}\sum_{i=1}^t \text{consistency}_i.
\end{align*}
A higher KI metric consistently reflects more accurate knowledge incorporated during generation, thus stronger knowledge capacity in the model.
We present more details about \texttt{gpt4o} input prompts and the calculation of the proposed metrics in Appendix~\ref{app:prompt_details}.

\begin{figure*}[t]
\centering
\includegraphics[width=\linewidth]{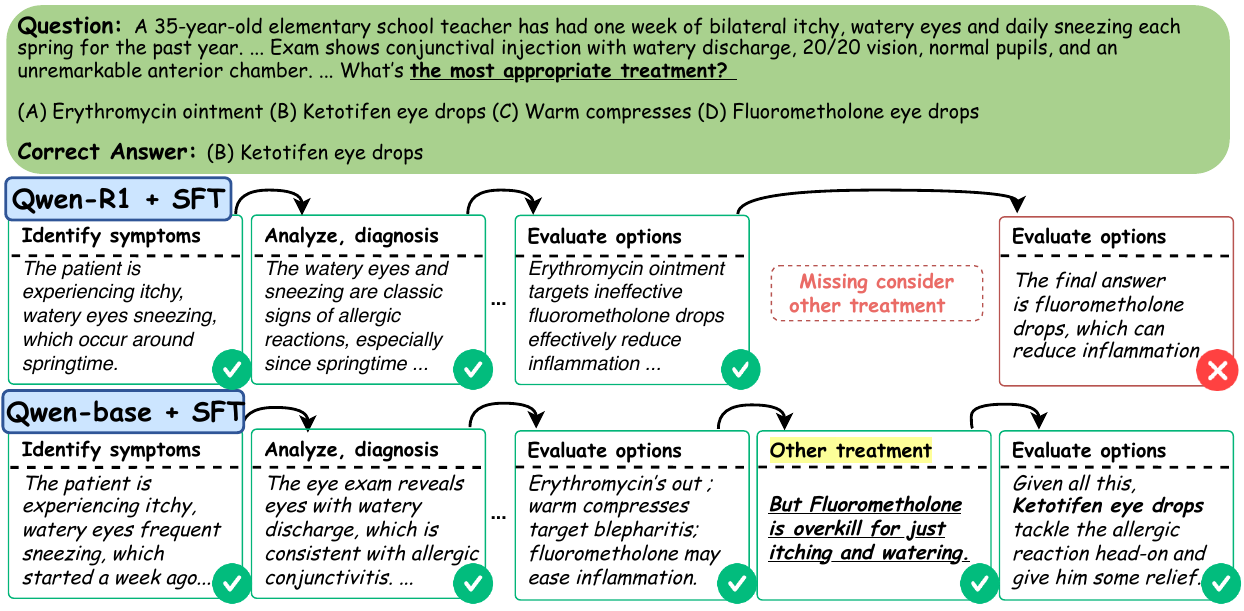} 
\caption{\textbf{Comparison of SFT Qwen-Base and Qwen-R1.} While the medical knowledge of all reasoning steps in Qwen-R1 + SFT is correct, the ignorance of considering more appropriate treatment for the specified disease results in an incorrect answer.}
\label{fig:R1_vs_base_case}
\end{figure*}

\section{Experiment Findings}

\subsection{Main Results}\label{sec:analysis_4.1}

% Table generated by Excel2LaTeX from sheet 'Table 1'
\begin{table}[htbp]
\small
  \centering
  \caption{\textbf{Comparison between base model and R1 distilled model.} After finetuning in the medical domain, base model demonstrates consistently superior performance across all metrics}
  \setlength{\tabcolsep}{0.8mm}
    \begin{tabular}{ccccccccc}
    \toprule
    \textbf{Base Model} & \textbf{SFT} & \textbf{RL} & \textbf{MedMCQA} & \textbf{MedQA} & \textbf{PubMedQA} & \textbf{MMLU-Pro} & \textbf{MedXpert} & \textbf{AVG} \\
    \midrule
    % \multicolumn{3}{l}{\textit{Metric: Accuracy (\%)}} & & & & low& \cellcolor[rgb]{ .816,  .918,  .851}$\rightarrow$ & \cellcolor[rgb]{ .388,  .745,  .482}high \\
    \multicolumn{3}{l}{\textit{Metric: Accuracy (\%)}} & & & & \multicolumn{3}{r}{\includegraphics[height=1.2em]{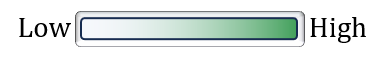}} \\
    \midrule
    \multicolumn{1}{c}{\multirow{2}[1]{*}{Qwen-R1}} & \ding{51}   & \ding{55}    & \cellcolor[rgb]{ .929,  .965,  .949}36.29 & \cellcolor[rgb]{ .941,  .973,  .961}35.69 & \cellcolor[rgb]{ .659,  .855,  .714}66.07 & \cellcolor[rgb]{ .816,  .918,  .851}41.54 & \cellcolor[rgb]{ .961,  .98,  .976}10.74 & \cellcolor[rgb]{ .851,  .933,  .882}38.07 \\
          & \ding{51}   & \ding{51}   & \cellcolor[rgb]{ .988,  .988,  1}33.20 & \cellcolor[rgb]{ .988,  .988,  1}32.60 & \cellcolor[rgb]{ .988,  .988,  1}50.70 & \cellcolor[rgb]{ .988,  .988,  1}30.49 & \cellcolor[rgb]{ .988,  .988,  1}10.40 & \cellcolor[rgb]{ .988,  .988,  1}31.48 \\
    \multicolumn{1}{c}{\multirow{2}[1]{*}{Qwen-Base}} & \ding{51}   & \ding{55}    & \cellcolor[rgb]{ .557,  .816,  .627}55.48 & \cellcolor[rgb]{ .529,  .804,  .604}62.61 & \cellcolor[rgb]{ .545,  .812,  .62}71.17 & \cellcolor[rgb]{ .537,  .808,  .612}59.11 & \cellcolor[rgb]{ .573,  .82,  .639}15.57 & \cellcolor[rgb]{ .541,  .808,  .616}52.79 \\
          & \ding{51}   & \ding{51}   & \cellcolor[rgb]{ .388,  .745,  .482}64.04 & \cellcolor[rgb]{ .388,  .745,  .482}71.56 & \cellcolor[rgb]{ .388,  .745,  .482}78.40 & \cellcolor[rgb]{ .388,  .745,  .482}68.27 & \cellcolor[rgb]{ .388,  .745,  .482}17.81 & \cellcolor[rgb]{ .388,  .745,  .482}60.02 \\
    \midrule
    % \multicolumn{3}{l}{\textit{Metric: Info Gain}} & & & &low& \cellcolor[rgb]{ .816,  .918,  .851}$\rightarrow$ & \cellcolor[rgb]{ .388,  .745,  .482}high \\
    \multicolumn{3}{l}{\textit{Metric: Info Gain}} & & & & \multicolumn{3}{r}{\includegraphics[height=1.2em]{figures/bar_low_to_high.jpg}} \\
    \midrule
    \multicolumn{1}{c}{\multirow{2}[1]{*}{Qwen-R1}} & \ding{51}   & \ding{55}    & \cellcolor[rgb]{ .988,  .988,  1}8.876 & \cellcolor[rgb]{ .663,  .859,  .718}0.139 & \cellcolor[rgb]{ .388,  .745,  .482}0.205 & \cellcolor[rgb]{ .988,  .988,  1}1.729 & \cellcolor[rgb]{ .988,  .988,  1}0.298 & \cellcolor[rgb]{ .988,  .988,  1}2.249 \\
          & \ding{51}   & \ding{51}   & \cellcolor[rgb]{ .545,  .808,  .616}9.202 & \cellcolor[rgb]{ .988,  .988,  1}0.113 & \cellcolor[rgb]{ .988,  .988,  1}0.183 & \cellcolor[rgb]{ .765,  .898,  .804}1.750 & \cellcolor[rgb]{ .988,  .988,  1}0.298 & \cellcolor[rgb]{ .627,  .843,  .69}2.309 \\
    \multicolumn{1}{c}{\multirow{2}[1]{*}{Qwen-Base}} & \ding{51}   & \ding{55}    & \cellcolor[rgb]{ .424,  .761,  .514}9.291 & \cellcolor[rgb]{ .439,  .769,  .529}0.157 & \cellcolor[rgb]{ .741,  .89,  .788}0.192 & \cellcolor[rgb]{ .388,  .745,  .482}1.785 & \cellcolor[rgb]{ .482,  .784,  .561}0.312 & \cellcolor[rgb]{ .396,  .749,  .49}2.347 \\
          & \ding{51}   & \ding{51}   & \cellcolor[rgb]{ .388,  .745,  .482}9.314 & \cellcolor[rgb]{ .388,  .745,  .482}0.161 & \cellcolor[rgb]{ .788,  .906,  .827}0.190 & \cellcolor[rgb]{ .631,  .847,  .694}1.762 & \cellcolor[rgb]{ .388,  .745,  .482}0.315 & \cellcolor[rgb]{ .388,  .745,  .482}2.348 \\
    \midrule
    % \multicolumn{4}{l}{\textit{Metric: Knowledge Index}} & & & low& \cellcolor[rgb]{ .816,  .918,  .851}$\rightarrow$ & \cellcolor[rgb]{ .388,  .745,  .482}high \\
    \multicolumn{3}{l}{\textit{Metric: Knowledge Index}} & & & & \multicolumn{3}{r}{\includegraphics[height=1.2em]{figures/bar_low_to_high.jpg}} \\
    \midrule
    \multicolumn{1}{c}{\multirow{2}[1]{*}{Qwen-R1}} & \ding{51}   & \ding{55}    & \cellcolor[rgb]{ .929,  .965,  .949}44.41 & \cellcolor[rgb]{ .831,  .925,  .867}61.97 & \cellcolor[rgb]{ .651,  .855,  .71}54.51 & \cellcolor[rgb]{ .671,  .863,  .729}69.32 & \cellcolor[rgb]{ .988,  .988,  1}52.24 & \cellcolor[rgb]{ .859,  .937,  .886}56.49 \\
          & \ding{51}   & \ding{51}   & \cellcolor[rgb]{ .988,  .988,  1}41.35 & \cellcolor[rgb]{ .988,  .988,  1}59.41 & \cellcolor[rgb]{ .988,  .988,  1}47.81 & \cellcolor[rgb]{ .988,  .988,  1}66.38 & \cellcolor[rgb]{ .804,  .914,  .839}56.48 & \cellcolor[rgb]{ .988,  .988,  1}54.29 \\
    \multicolumn{1}{c}{\multirow{2}[1]{*}{Qwen-Base}} & \ding{51}   & \ding{55}    & \cellcolor[rgb]{ .667,  .859,  .722}57.92 & \cellcolor[rgb]{ .482,  .784,  .565}67.65 & \cellcolor[rgb]{ .388,  .745,  .482}59.69 & \cellcolor[rgb]{ .765,  .898,  .808}68.48 & \cellcolor[rgb]{ .506,  .792,  .584}63.24 & \cellcolor[rgb]{ .439,  .769,  .529}63.40 \\
          & \ding{51}   & \ding{51}   & \cellcolor[rgb]{ .639,  .847,  .698}59.23 & \cellcolor[rgb]{ .388,  .745,  .482}69.17 & \cellcolor[rgb]{ .631,  .843,  .69}54.95 & \cellcolor[rgb]{ .388,  .745,  .482}71.93 & \cellcolor[rgb]{ .388,  .745,  .482}65.87 & \cellcolor[rgb]{ .388,  .745,  .482}64.23 \\
    \bottomrule
    \end{tabular}%
  \label{tab:Qwen_vs_R1}%
\end{table}%

We first focus on probing the performance of different models on the general accuracy or the knowledge and reasoning aspects in the medical domain.
We select two types of base models for training: Qwen-Base (Qwen-2.5-7B-base) and Qwen-R1 (DeepSeek-R1 distilled Qwen2.5-7B). 
Subsequently, we follow the conventional paradigm of deploying SFT and SFT + RL~\citep{deepseek_r1} on LLMs under the same setting and present results in Table~\ref{tab:Qwen_vs_R1}.

\begin{wrapfigure}{r}{0.45\linewidth}
    \centering
    \includegraphics[width=.93\linewidth]{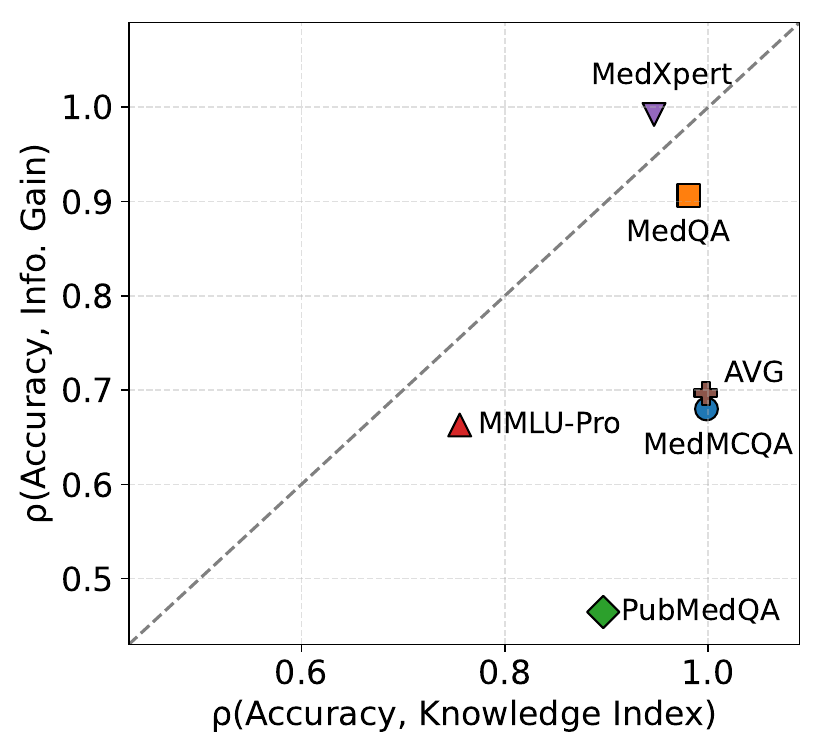}
    \caption{\textbf{Correlations between the proposed two metrics and accuracy.} Different tasks require different levels of knowledge and/or reasoning capabilities in LLMs.}
    \label{fig:metrics_corr}
\end{wrapfigure}

\paragraph{Trained Qwen-Base Outperforms its R1-distilled Counterpart.}
From the overall accuracy results, we observe that Qwen-Base consistently outperforms the R1-distilled variant across the evaluated benchmarks, whether using SFT alone or in combination with subsequent RL.
For instance, Qwen-Base with only SFT witnesses a 14.7\% average accuracy improvement over the distilled model (52.79\% \textit{vs.} 38.07\%). With the addition of RL, the gap widens further with a 22.6\% increase in average accuracy (54.06\% \textit{vs.} 31.48\%).
This interesting performance difference may stem from a domain shift across training stages. Since Qwen-R1 was primarily trained on R1-generated texts focused on math and code~\cite{deepseek_r1}, the medical domain knowledge introduced during our post-training could conflict with its prior representations, thereby undermining its performance in medical tasks.

To illustrate the advantage of the trained Qwen-Base over Qwen-R1 more intuitively, we include a case in~\cref{fig:R1_vs_base_case}, where Qwen-R1 + SFT fails to produce the correct answer.
In this example, although the model demonstrates sound reasoning, it selects Fluorometholone without evaluating safer alternatives. Ketotifen, with fewer side effects, would have been the better choice with less side effect---highlighting the importance of nuanced decision-making in medicine, unlike the single-solution nature of math problems.\looseness=-1

% \vspace{-0.6cm}

\paragraph{Knowledge and Reasoning Should be Two Distinct Evaluation Aspects for LLMs.}
When turning our focus on reasoning and knowledge as separate evaluation dimensions, we find that these aspects should be treated independently. 
While the reasoning abilities of the two Qwen-Base variants remain comparable, the RL-ed Qwen-Base demonstrates slightly better medical knowledge according to our KI metric (64.2 \textit{vs.} 63.4).
Similarly, although the RL-enhanced Qwen-R1 shows a minor improvement over its SFT-only version in general performance, it underperforms in knowledge evaluation, trailing by 2.2 points in the KI metric (54.3).
This pattern is also observed in specific datasets like MMLU-Pro and MedMCQA, hence supporting our argument for a clear separation between reasoning and knowledge evaluation.

\paragraph{The Challenging Nature of Different Benchmarks may Inherit from Different Aspects.} 
% \haoqin{TBD}
Separately evaluating knowledge and reasoning raises the question of which contributes more to the final accuracy. 
To explore this, we compute the correlation scores between each capability and the accuracy of the task in Figure~\ref{fig:metrics_corr} with the values on the x axis and on the y axis representing accuracy and KI, accuracy and Info. Gain correlations, respectively.
Our findings suggest that benchmark difficulty often hinges more on specific aspects. In PubMedQA, knowledge dominates, with KI correlating more strongly with accuracy than InfoGain (0.897 \textit{vs.} 0.465).
By contrast, tasks like MedQA and MedXpert depend on both aspects, as shown by correlations above 0.9 for both metrics.
The figure also implies that most medical benchmarks are more knowledge-intensive than reasoning-driven, as all correlation points except one fall below the x=y line. 
This line represents parity between the correlation of accuracy with knowledge and with reasoning. 
% The fact that only one point (MedXpert) lies above this line suggests that, for most benchmarks, knowledge is more strongly correlated with performance than reasoning ability.
This trend is reinforced by the average scores across all tasks, where KI shows a striking 0.998 correlation with the final accuracy---0.3 higher than that of reasoning.
% For instance, when comparing two Qwen-Base variants on MMLU-Pro, the SFT model outperforms the SFT+RL variant in the knowledge dimension (71.93 $>$ 68.48), while slightly underperforming in reasoning (1.76 $<$ 1.79), yet still achieves higher overall accuracy (60.97 $>$ 59.11).

\subsection{SFT \textit{vs.} RL in Medical and Math Domains}\label{sec:analysis_4.2}

\begin{figure}[t!]
\hfill
    \includegraphics[width=0.99\linewidth]{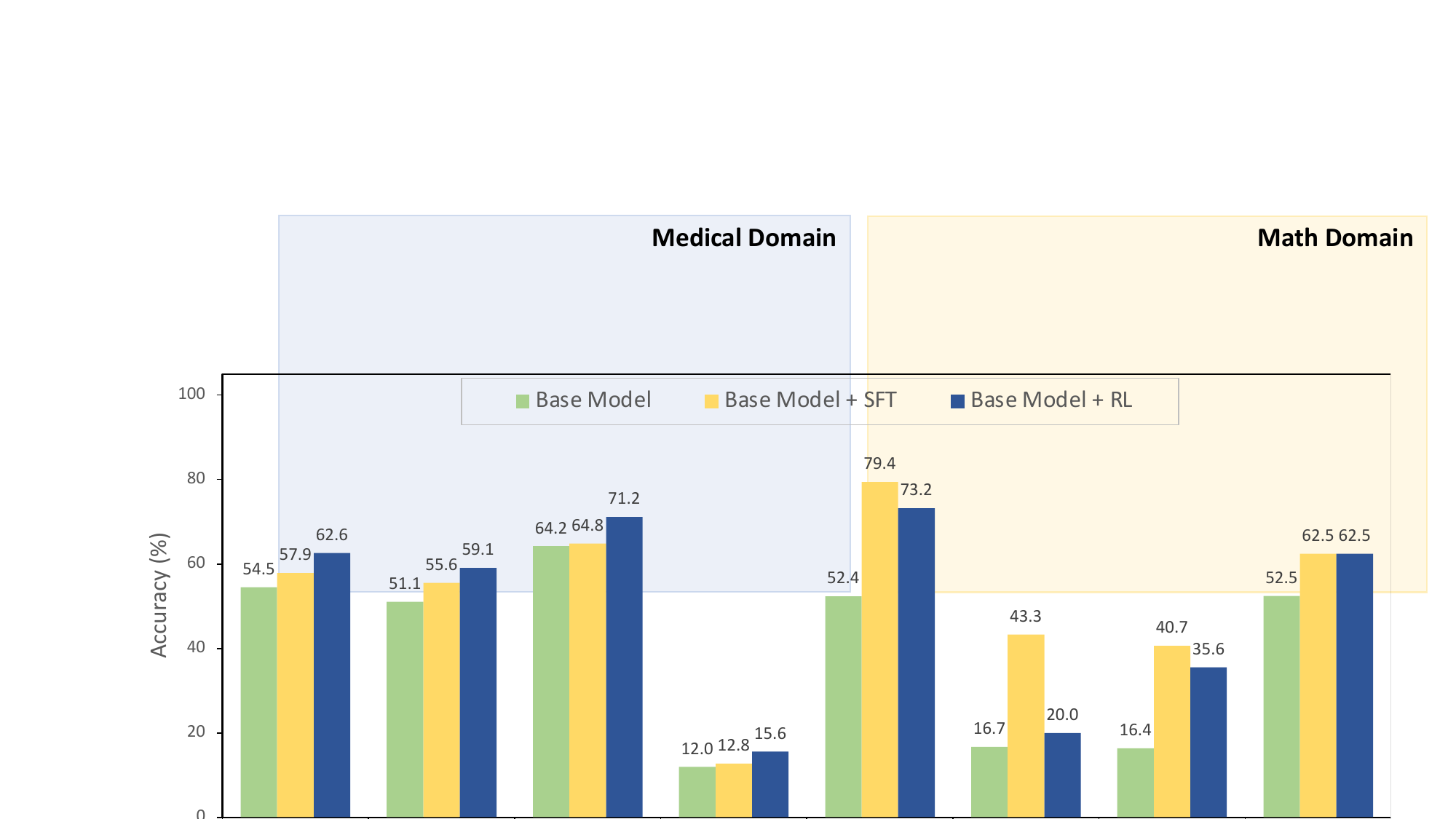}
    \caption{\textbf{Comparison between medical and math domain.} (a) In mathematics, RL enhances accuracy most, whereas in the medical field, SFT provides a greater enhancement in overall accuracy; (b) Across both fields, RL is more adept at boosting information gain, whereas SFT results in a reduction of information gain; (c) Within the medical field, SFT achieves the pinnacle of the knowledge index.}
    \label{fig:Math_vs_Med}
\end{figure}

Motivated by recent researches focusing on exploring the roles of SFT and RL in training reasoning models, we take a step forward on the popular training paradigm of ``SFT then RL'' and leverage SFT or RL separately on medical and math tasks. In detail, for the medical capability we employ Qwen2.5-7B-Base~\citep{qwen25_base} on the medical-o1 data~\citep{chen2024huatuogpt}, while in the math domain, we take Qwen2.5-7B-math~\citep{qwen25_math} as the base model trained on RLHFlow data~\citep{online_dpo_r1,dong2024rlhfworkflow}. We only present the knowledge index metric for medical benchmarks, as the knowledge base of math.

\paragraph{SFT Usually Provides Knowledge with Reasoning Compromised, RL Boosts Both.}
Figure(c)~\ref{fig:Math_vs_Med} shows that in the medical domain, both RL and SFT encourage the model to incorporate more domain-specific knowledge. 
SFT consistently yields a greater boost than RL, with average absolute gains in the KI metric of 13.69 and 10.53 over the base model, respectively (base 49.71 $<$ RL-ed 60.24 $<$ SFT-ed 63.40).
By contrast, the reasoning ability of models follows an opposite trend in Figure~\ref{fig:Math_vs_Med}(b): RL consistently enhances reasoning across both medical and math domains (0.39 \textit{vs.} 0.42 in medical, 0.13 \textit{vs.} 0.15 in math).
SFT, however, appears to hinder reasoning, leading to notable drops of 0.17 and 0.04 in the medical and math domains, respectively (0.22 \textit{vs.} 0.39 in medical, 0.09 \textit{vs.} 0.13 in math). This drop accounts for an average reduction of 37.1\% in the original Info Gain scores.

These findings suggest that while RL promotes a more balanced improvement in both knowledge and reasoning, SFT tends to favor domain knowledge (\textit{e.g.}, medical facts) at the cost of undermining the model’s inherent reasoning ability, consistent with previous observations~\citep{chu2025sft,chen2025sft}.

\paragraph{Medical Problems are Likely to Require More Knowledge While Math Needs Enhanced Reasoning.}
Final accuracy on medical tasks, shown in Figure~\ref{fig:Math_vs_Med}(a), indicates that the SFT-ed model outperforms both its RL-ed and base counterparts, with average gains of 4.6\% and 6.2\%, respectively (SFT-ed 49.8\% $>$ RL-ed 49.8\% $>$ base 48.2\%).
However, in the math domain, the RL-ed variant leads to the highest accuracy, surpassing both the base and SFT-ed models (average RL-ed 61.7\% $>$ SFT-ed 51.9\% $>$ base 40.5\%).
Given the earlier evidence that SFT enhances domain knowledge while RL improves reasoning, we conclude that, unlike reasoning-intensive tasks such as math problems, challenging medical tasks benefit more from additional domain knowledge, resulting in higher task accuracy.

\begin{figure}[t]
    \includegraphics[width=0.99\linewidth]{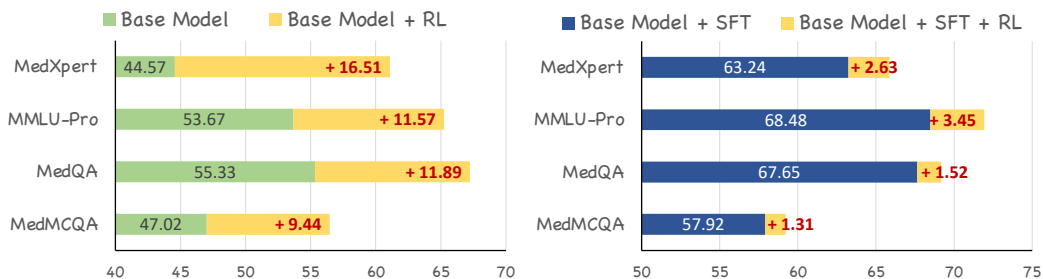}
    \caption{\textbf{RL improves knowledge Index.} In the medical domain, applying RL to base model before or post SFT consistently improves the knowlegde index.}
    \label{fig:RL_Knowledge}
\end{figure}

\paragraph{RL can Improve Model's Medical Knowledge Correctness.}
While RL is widely recognized for enhancing medical reasoning abilities~\citep{chen2024huatuogpt}, it can also improve the knowledge presented within reasoning trajectories, as discussed earlier.
To explore this effect, we analyze Qwen-Base with either RL alone or SFT followed by RL, as shown in Figure~\ref{fig:RL_Knowledge}.
The results indicate that RL boosts the knowledge index metric across different medical tasks, with notable gains of 12.4 points (RL alone) and 2.2 points (SFT+RL) over their respective RL-free baselines.
This is particularly interesting given that RL introduces little new knowledge to the model~\citep{chu2025sft,chen2025sft}.

\begin{figure*}[t]
\centering
\includegraphics[width=\linewidth]{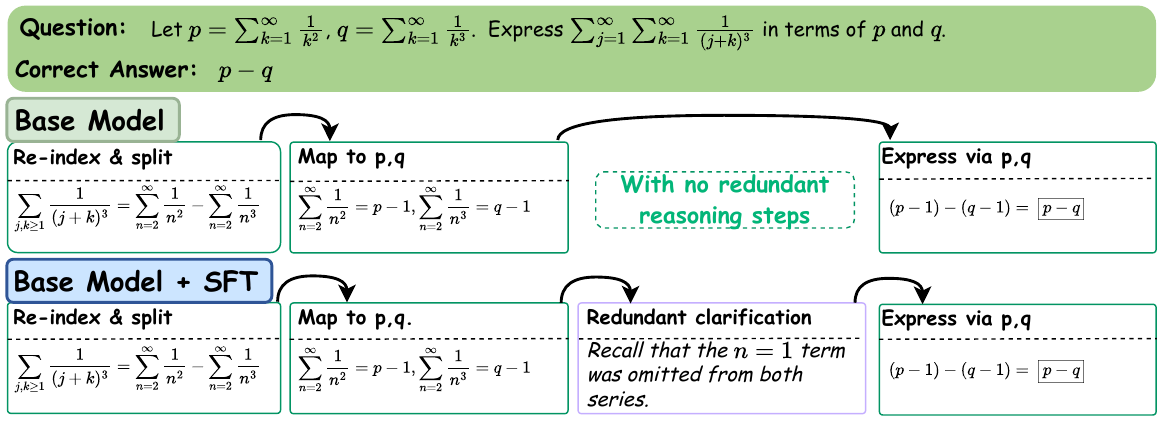} 
\caption{\textbf{Comparison of Models w and wo SFT} We compare both factual grounding and reasoning efficiency. SFT adds redundant reasoning steps, reducing per-step information gain and inference efficiency.}
\label{fig:base_vs_SFT_redundant}
\end{figure*}

\begin{figure*}[t]
\centering
\includegraphics[width=\linewidth]{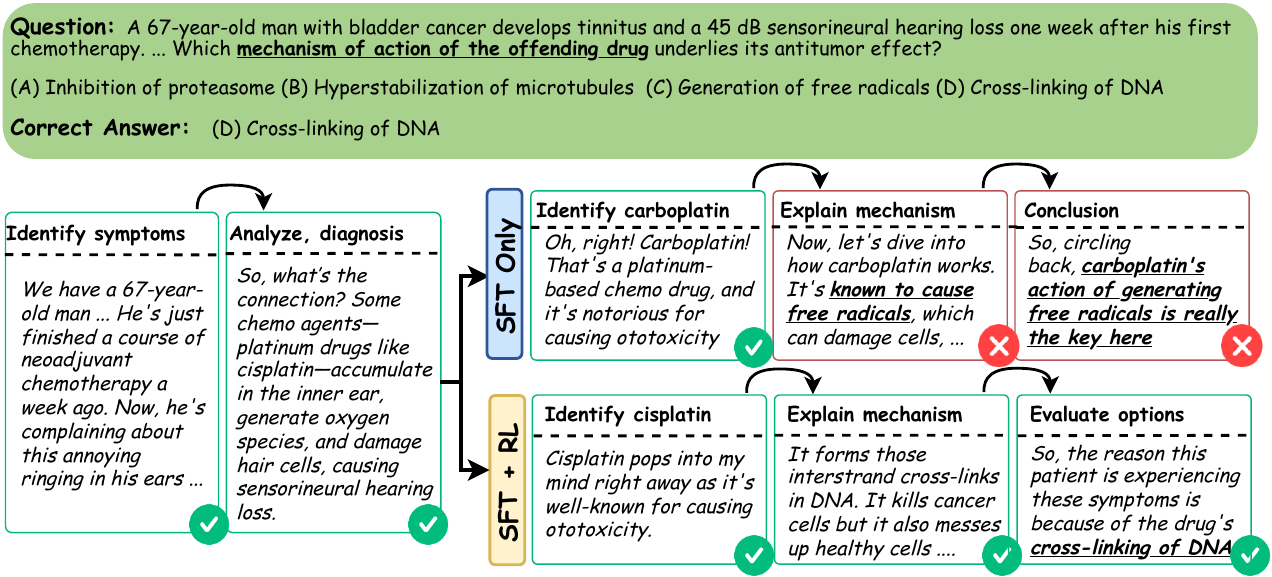} 
\caption{\textbf{Comparison of models w/ and w/o RL.} We compare the knowledge correctness of the models' reasoning steps. RL improves the knowledge correctness by guiding the model to select reasoning paths that have fewer knowledge errors.}
\vspace{-.5em}
\label{fig:SFT_vs_RL_case}
\end{figure*}

To further investigate, we conduct a case study revealing that RL enhances knowledge metrics by guiding the model to discard reasoning paths containing incorrect knowledge, rather than adding new facts.
We refer to this effect as enhanced knowledge correctness.
As shown in~\cref{fig:SFT_vs_RL_case}, both the SFT-only and SFT+RL models begin similarly, correctly linking symptoms to chemotherapy-induced hearing loss.
The key challenge lies in identifying which chemotherapy drugs are ototoxic. While both cisplatin and carboplatin fit this criterion and share a DNA cross-linking mechanism, the SFT-only model incorrectly selects carboplatin and misattributes its mechanism to free radical generation, leading to an incorrect answer.
In contrast, the SFT+RL model correctly selects cisplatin, avoiding the knowledge error.

\section{Discussion and Conclusion}
\label{sec:conclusion_discussion}
\paragraph{Limitations.}
Our evaluation framework explicitly separates knowledge from reasoning steps, though this initial study has its limitations.
In particular, all experiments were conducted on the Qwen family of 7B-parameter models. Nevertheless, the two tested variants (Qwen-R1 and Qwen-Base) reflect strong generalization, robustness, and high-capacity. 
We therefore consider the findings reasonably convincing within this scope.
We acknowledge that evaluating a broader range of models is necessary to assess the generality of our conclusions.
Although our benchmark focuses on mathematics and medicine, these domains were chosen deliberately: mathematics serves as a conventional testbed for reasoning in LLMs~\citep{sessler2024benchmarking}, while medicine represents a high-stakes domain where factual accuracy and domain knowledge are essential for ensuring safety and trust~\citep{nori2023capabilities}.

\paragraph{Broader Impact.}
Looking ahead, we believe our evaluation framework can be extended to other structured reasoning domains. In legal tasks, for instance, the IRAC\cite{servantez2024chain} (Issue, Rule, Application, Conclusion) structure offers a natural alignment with stepwise evaluation for knowledge and reasoning decomposition~\citep{servantez2024chain}. 
A model could be prompted to identify the issue, retrieve relevant laws (knowledge), apply them to the case (reasoning), and conclude—allowing our framework to assess performance at each stage.
Early work on legal reasoning with LLMs (\textit{e.g.}, LegalBench~\citep{guha2023legalbench}, Chain of Logic prompting~\citep{servantez2024chain}) highlights the value of explicit multi-step reasoning.
Similarly, financial and economic tasks\cite{yu2023temporal} involve structured reasoning, often combining domain knowledge with time-series data. Recent studies show LLMs can integrate textual knowledge with historical data to make forecasts~\citep{yu2023temporal}. A typical financial problem may involve summarizing trends (knowledge), applying economic models (reasoning), and predicting outcomes.
Although processing raw numerical data remains challenging, promising results suggest that LLMs can perform sequential reasoning over such inputs.
Adapting our knowledge-\textit{vs.}-reasoning framework to these domains, despite their structural differences, offers a promising path to better understand and improve LLM reasoning in complex, high-stakes settings.

\paragraph{Conclusion.}
In this paper, we put our primary focus on decomposing the reasoning paths of LLMs into knowledge and reasoning components to better understand their contributions to final performance. 
Through our proposed evaluation framework and analysis of Qwen models trained with SFT and RL, we demonstrate that these two capacities are not only distinct but also unequally demanded across domains. 
Our findings reveal that SFT boosts factual knowledge, especially in knowledge-intensive domains like medicine, albeit at the cost of reasoning clarity, while RL improves reasoning quality by pruning incorrect knowledge. 
This decomposition offers a more transparent lens into LLM decision-making and paves the way for targeted improvements in domain-specific reasoning tasks.

\section{Acknowledgement}
This work was partially funded by an unrestricted gift from Google.
We thank the Microsoft Accelerate Foundation Models Research Program for supporting our computing needs.

\bibliographystyle{plain}
\bibliography{neurips_2025}

%%% END INSTRUCTIONS %%%
\newpage
\appendix

\section*{Technical Appendices and Supplementary Material}
Technical appendices with additional results, figures, graphs and proofs may be submitted with the paper submission before the full submission deadline (see above), or as a separate PDF in the ZIP file below before the supplementary material deadline. There is no page limit for the technical appendices.

% \subsection{Reasoning Decompose Example}

\section{Result of ROSCOE Metric}
% Table generated by Excel2LaTeX from sheet 'Other Metrics'
\begin{table}[htbp]
  \centering
  \caption{\textbf{Evaluation using ROSCOE-SA metrics.} We employ ROSCOE semantic alignment metrics (ROSCOE-SA) including Faithfulness-Step and Informativeness-Step. Despite the knowledge index and information gain differing among the base model, SFT-ed model, and RL-ed model, these metrics yield comparable evaluations for each.}
  \setlength{\tabcolsep}{0.9mm}
    \begin{tabular}{cccccccc}
    \toprule
    \textbf{Base Model} & \textbf{SFT} & \textbf{RL} & \textbf{MedMCQA} & \textbf{MedQA} & \textbf{PubMedQA} & \textbf{MMLU-Pro} & \textbf{AVG} \\
    \midrule
    \multicolumn{4}{l}{\textit{Metric: ROSCOE-SA Faithfulness-Step}} &       &    &  \\
    \midrule
    \multicolumn{1}{c}{\multirow{3}[2]{*}{Qwen-Base}} & \ding{55}    & \ding{55}    & 0.744 & 0.776 & 0.842 & 0.762 & 0.781 \\
          & \ding{51}   & \ding{55}    & 0.742 & 0.777 & 0.843 & 0.760 & 0.781 \\
          & \ding{51}   & \ding{51}   & 0.742 & 0.778 & 0.842 & 0.760 & 0.781 \\
    \midrule
    \multicolumn{4}{l}{\textit{Metric: ROSCOE-SA Informativeness-Step}}     &       &       &  \\
    \midrule
    \multicolumn{1}{c}{\multirow{3}[2]{*}{Qwen-Base}} & \ding{55}    & \ding{55}    & 0.776 & 0.772 & 0.801 & 0.776 & 0.781 \\
          & \ding{51}   & \ding{55}    & 0.774 & 0.773 & 0.802 & 0.775 & 0.781 \\
          & \ding{51}   & \ding{51}   & 0.775 & 0.774 & 0.802 & 0.775 & 0.782 \\
    \bottomrule
    \end{tabular}%
  \label{tab:ROSCOE_res}%
\end{table}%

\section{Experiment Details}
\label{app:training_details}

\subsection{Experimental Settings}

\subparagraph{Models.}  
All experiments in both the medical and math domains are initialized from the universal 7B-parameter base models:
\begin{itemize}
  \item \textbf{Qwen2.5‐7B}~\citep{qwen25_base} is an open-weigh language model that takes leading positions in a wide range of tasks. It is pretrained on a large‑scale multilingual, multi-domain corpus.
  \item \textbf{DeepSeek‐R1‐Distill‐Qwen‐7B}~\citep{deepseek_r1} is a distilled variant of Qwen2.5-7B model. Being trained with supervised fine‐tuning (SFT) on DeepSeek‐R1 distilled data, this model develops inherent thinking abilities and presents superior performance especially on math and coding tasks.
\end{itemize}
We select these two models as baselines due to: (1) the strong generalization and robustness of Qwen-7B across domains; (2) their open-source nature, enabling in-depth exploration of training, evaluation, and architecture; and (3) their compatibility for fair comparison on the impact of post-training thinking patterns.

\paragraph{Training Datasets and Details.}
For the medical domain, we fine-tune two LLMs using either SFT or RL on the corresponding data split from \textit{medical‐o1}~\citep{chen2024huatuogpt}.
For models in the math domain, we employ the SFT or RL pipeline from RLHFlow~\citep{online_dpo_r1} to train the Qwen models, resulting in four model variants on two Qwen models with SFT or RL. 
We provide detailed medical training data below:
\begin{itemize}
  \item \textbf{Supervised Fine‑Tuning (SFT):}
        \href{https://huggingface.co/datasets/FreedomIntelligence/medical-o1-reasoning-SFT}{\texttt{medical‑o1‑reasoning‑SFT}} contains \(\sim\!40{,}000\) physician‑level question–answer pairs.  
        Each record comprises  
        \emph{(i)} an \verb|instruction| field (clinical prompt),  
        \emph{(ii)} a \verb|complex_cot| sequence with comprehensive chain‑of‑thought generated by a multi-round data pipeline, and  
        \emph{(iii)} a concise \verb|response|.  

  \item \textbf{Reinforcement Learning (RL):}
        \href{https://huggingface.co/datasets/FreedomIntelligence/medical-o1-verifiable-problem}{\texttt{medical‑o1‑verifiable‑problem}} offers the same \(\sim\!40{,}000\) clinical cases but retains only the \verb|question| and a single \verb|ground_truth| answer.  
        The absence of chain‑of‑thought makes each item a \emph{verifiable prompt}: model outputs are scored by an external medical verifier model with a sparse, sign‑based reward (+1 for correct, 0 for incorrect).
\end{itemize}
The training processes involved reinforcement learning are conducted on 4 NVIDIA H100 GPUs, costing 12 hours for training each model. SFT are conducted on 8 NVIDIA A5000 GPUs, costing 28 hours for training each model.

\textbf{Evaluation Datasets}
We evaluate our models on two representative domains---\emph{medicine} and \emph{mathematics}---using the public benchmarks listed below.

\begin{description}

  % ------------------ Medical ------------------ %
  \item[\textbf{Medical}]%
  \begin{itemize}
    \item \href{https://medmcqa.github.io/}{MedMCQA}~\citep{pal2022medmcqa} -- $\sim$194,000 four‑option MCQs drawn from India’s AIIMS / NEET‑PG entrance exams, covering 21 subjects and 2 400 + topics.
    \item \href{https://paperswithcode.com/dataset/medqa-usmle}{MedQA‑USMLE}~\citep{jin2020medqa} -- $\sim$12,000 multiple choice questions patterned after USMLE Steps 1–3 in both English and Chinese.
    \item \href{https://pubmedqa.github.io/}{PubMedQA}~\citep{jin2019pubmedqa} -- $\sim$61,000 yes / no / uncertain questions, each paired with a PubMed abstract for context.
    \item \href{https://huggingface.co/datasets/TIGER-Lab/MMLU-Pro}{MMLU‑Pro (Medical)}~\citep{wang2024mmlupro} -- $\sim$1,050 ten‑option items from the professional‑medicine slice of the MMLU Pro suite.
    \item \href{https://arxiv.org/abs/2501.18362}{MedXpertQA}~\citep{zuo2025medxpert} -- 4 ,460 expert-level free response questions spanning 17 specialties and 11 body systems.
  \end{itemize}

  % ---------------- Mathematics ---------------- %
  \item[\textbf{Mathematics}]%
  \begin{itemize}
    \item \href{https://artofproblemsolving.com/wiki/index.php/AIME_Problems_and_Solutions}{AIME 2024}\cite{aime2024} -- 15 open‑response problems from the 2024 American Invitational Mathematics Examination.
    \item \href{https://www.vals.ai/benchmarks/math500-03-24-2025}{MATH500}\cite{vals2025math500} -- 500 tasks sampled from the original MATH benchmark, covering algebra, geometry, number theory, and combinatorics.
    \item \href{https://huggingface.co/datasets/hendrycks/competition_math}{AMC (10 \& 12)}\cite{hendrycks2021math} -- multi-year collections of 25-item multiple choice tests widely used in high‑school math studies.
    \item \href{https://github.com/EleutherAI/lm-evaluation-harness/tree/main/lm_eval/tasks/minerva_math}{Minerva‑Math}\cite{lewkowycz2022minerva} -- quantitative reasoning subset introduced with Google's Minerva model, focusing on chain-of-thought solutions.
    \item \href{https://www.maa.org/math-competitions}{USAMO 2025 / Olympiad}\cite{usamo2025} -- six proof‑based questions from the 2025 USA Mathematical Olympiad for rigorous derivation evaluation.
  \end{itemize}

\end{description}

\begin{table}[htbp]
  \centering
  \small
  \begin{tabular}{@{}llc@{}}
    \toprule
    \textbf{Stage} & \textbf{Key Parameter} & \textbf{Value} \\
    \midrule
    \multirow{7}{*}{Supervised FT} 
      & Epochs & 3 \\
      & Effective batch size & 128 sequences \\[-2pt]
      & Optimizer & AdamW\cite{loshchilov2019decoupled} ($\beta_1{=}0.9,\ \beta_2{=}0.999$, wd 0.01) \\
      & LR schedule & linear warm-up $\rightarrow$ cosine decay, $\text{LR}_{\max}=3{\times}10^{-5}$ \\
      & Max sequence length & 1 024 tokens \\
      & Precision & \texttt{bf16} \\
      & Parallel engine & DeepSpeed ZeRO-3 (param\&opt off-load CPU) \\
    \midrule
    \multirow{10}{*}{RL (PPO)} 
      & Total episodes & 20 000 \\
      & PPO epochs / update & 3 \\
      & Mini-batches / epoch & 1 \\
      & Rollout batch size & 4 prompts (per forward pass) \\
      & Clip range $\epsilon$ & 0.2 (implicit default) \\
      & KL penalty $\lambda_{\text{KL}}$ & 0.03 \\
      & Target KL & 6.0 (early LR reduce) \\
      & Actor LR & $5{\times}10^{-7}$ \\
      & Warm-up ratio & 0.05 \\
      & Grad accumulation & 16 (effective bsz = 64) \\
    \bottomrule
  \end{tabular}
  \caption{Principal hyper-parameters for supervised fine-tuning (SFT) and PPO\cite{schulman2017ppo} reinforcement learning on the \textit{medical-o1} corpus. Both stages employ bf16 mixed precision and DeepSpeed ZeRO-3\cite{rajbhandari2020zero} on a single 8-GPU node.}
  \label{tab:medical_hparams}
\end{table}

\subsection{Prompts for Metric Calculation.}
\label{app:prompt_details}
This section provides the detailed prompts employed in the evaluation pipeline (\cref{sec:method}).

\begin{figure*}[htbp]
\centering
\includegraphics[width=\linewidth]{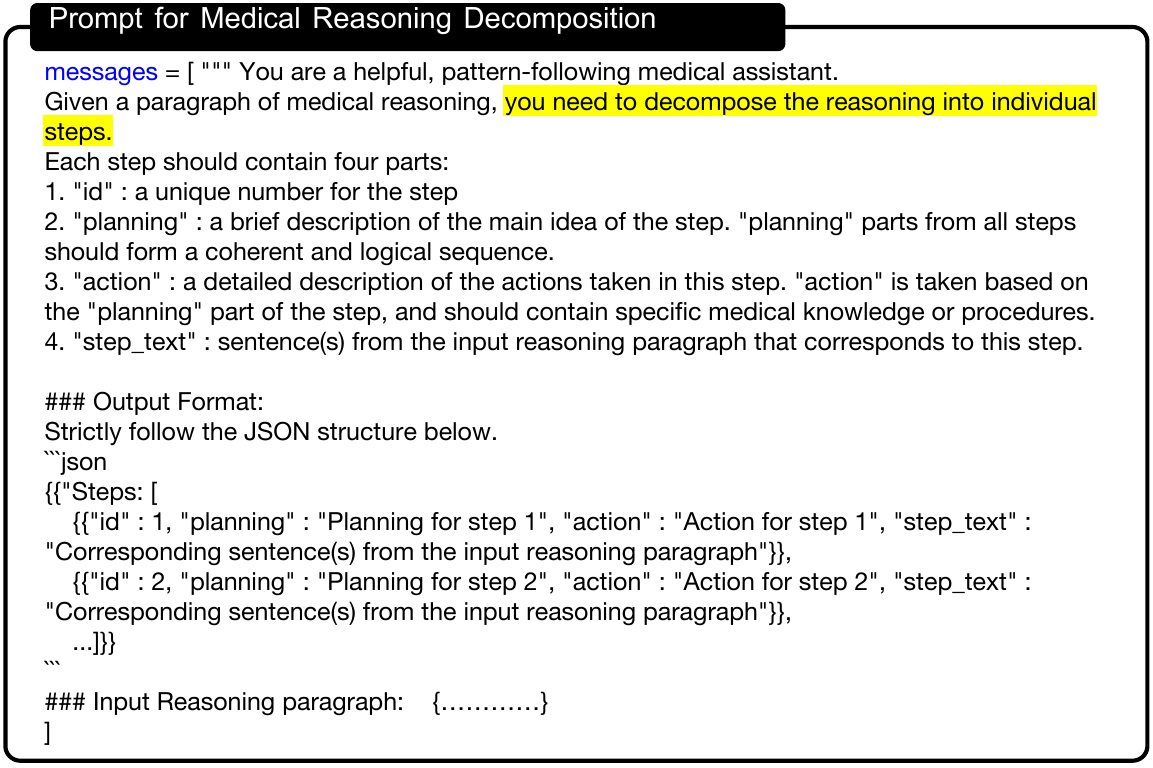} 
\caption{\textbf{Medical Reasoning Decomposition Prompt} Full prompt employed to decompose the model's reasoning into reasoning steps using \texttt{gpt4o}.}
\label{fig:Medical_Reasoning_Decomposition}
\end{figure*}

\begin{figure*}[htbp]
\centering
\includegraphics[width=\linewidth]{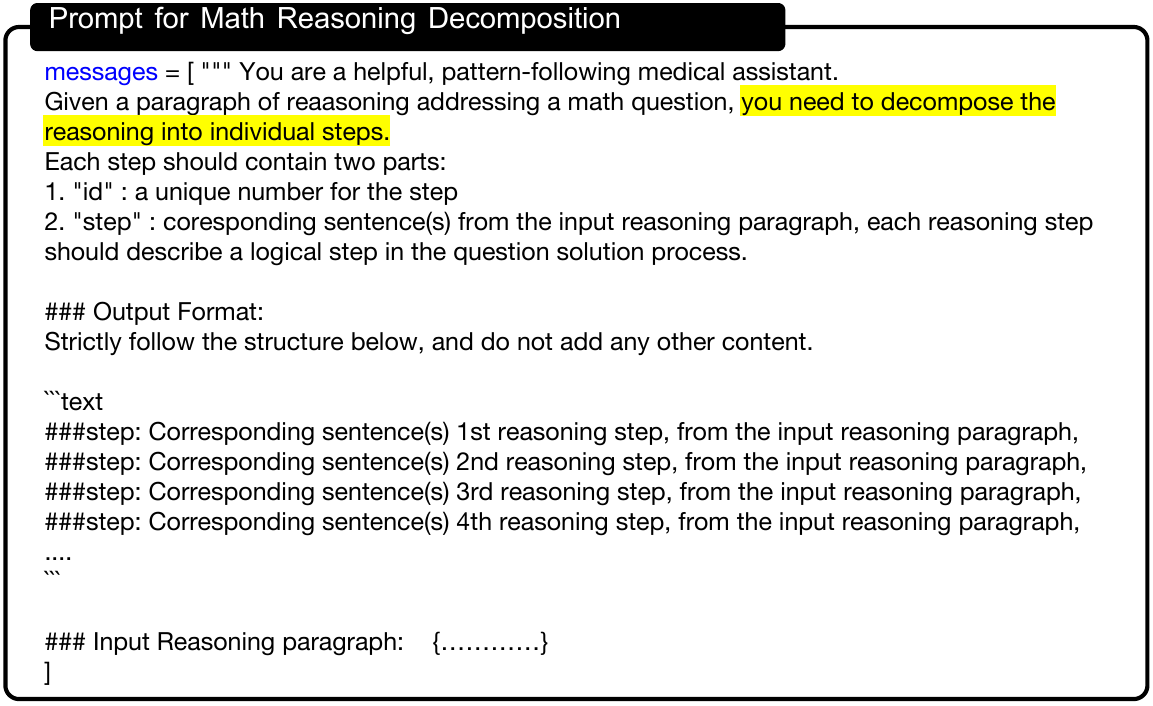} 
\caption{\textbf{Math Reasoning Decomposition Prompt} Full prompt employed to decompose the model's reasoning into reasoning steps using \texttt{gpt4o}.}
\label{fig:Math_Reasoning_Decomposition}
\end{figure*}

\begin{figure*}[htbp]
\centering
\includegraphics[width=\linewidth]{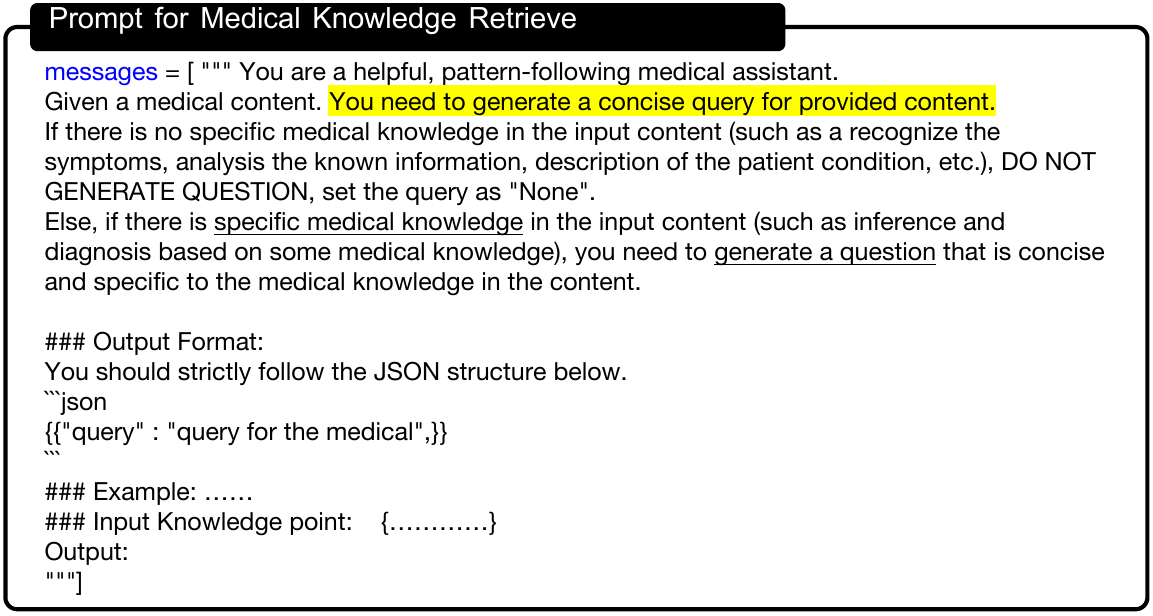} 
\caption{\textbf{Medical Knowledge Retrieve Prompt} Full prompt employed to identify the knowledge pertinent to the reasoning step using \texttt{gpt4o}.}
\label{fig:Medical_Knowledge_Retrieve}
\end{figure*}

\begin{figure*}[htbp]
\centering
\includegraphics[width=\linewidth]{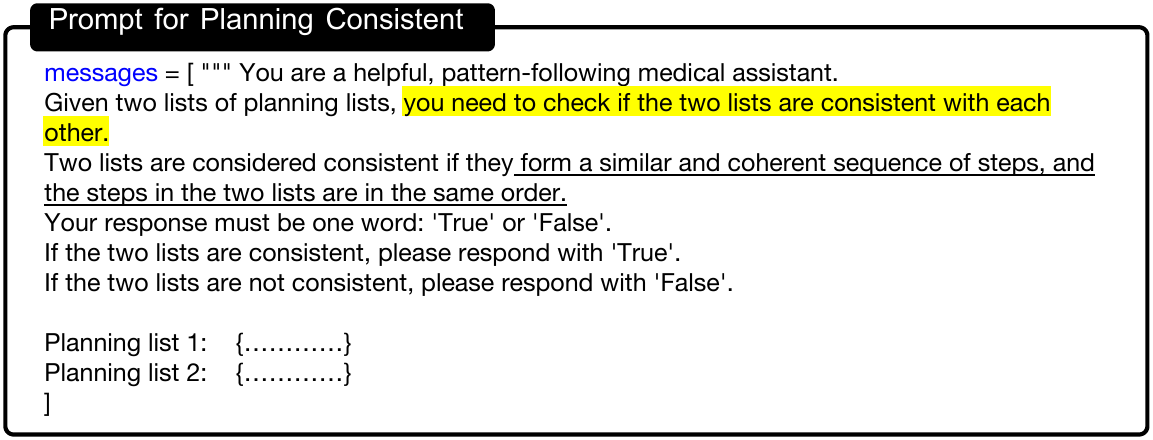} 
\caption{\textbf{Planning Consistency Evaluation Prompt} Full prompt employed to evaluate whether the reasoning step is consistent with the retrieved facts using \texttt{gpt4o}.}
\label{fig:Planning_Consistent}
\end{figure*}

\section{More experiment results}
% Table generated by Excel2LaTeX from sheet 'Table 4'
\begin{table}[htbp]
  \centering
  \setlength{\tabcolsep}{0.9mm}
  \caption{\textbf{Comparison between different SFT data filtering strategies.} Quality filtering achieves comparable performance in terms of accuracy and information gain as full data, while reducing the training cost.}
    \begin{tabular}{ccccccc}
    \toprule
    \textbf{Data Filtering} & \multicolumn{1}{c}{\textbf{MedMCQA}} & \multicolumn{1}{c}{\textbf{MedQA}} & \multicolumn{1}{c}{\textbf{PubMedQA}} & \multicolumn{1}{c}{\textbf{MMLU-Pro}} & \multicolumn{1}{c}{\textbf{MedXpert}} & \multicolumn{1}{c}{\textbf{Avg}} \\
    \midrule
    \multicolumn{7}{l}{\textit{Metric: Accuracy}} \\
    \midrule
    None  & \cellcolor[rgb]{ .388,  .745,  .482}55.48 & \cellcolor[rgb]{ .467,  .776,  .549}62.61 & \cellcolor[rgb]{ .388,  .745,  .482}71.17 & \cellcolor[rgb]{ .388,  .745,  .482}59.11 & \cellcolor[rgb]{ .388,  .745,  .482}15.57 & \cellcolor[rgb]{ .388,  .745,  .482}52.79 \\
    Difficulty & \cellcolor[rgb]{ .988,  .988,  1}52.43 & \cellcolor[rgb]{ .988,  .988,  1}57.63 & \cellcolor[rgb]{ .988,  .988,  1}65.10 & \cellcolor[rgb]{ .988,  .988,  1}57.39 & \cellcolor[rgb]{ .988,  .988,  1}13.83 & \cellcolor[rgb]{ .988,  .988,  1}49.28 \\
    Quality & \cellcolor[rgb]{ .42,  .761,  .51}55.33 & \cellcolor[rgb]{ .388,  .745,  .482}63.34 & \cellcolor[rgb]{ .404,  .753,  .494}71.03 & \cellcolor[rgb]{ .533,  .804,  .608}58.70 & \cellcolor[rgb]{ .69,  .867,  .741}14.70 & \cellcolor[rgb]{ .42,  .757,  .51}52.62 \\
    \midrule
    \multicolumn{7}{l}{\textit{Metric: Info Gain}} \\
    \midrule
    None  & \cellcolor[rgb]{ .388,  .745,  .482}9.291 & \cellcolor[rgb]{ .388,  .745,  .482}0.157 & \cellcolor[rgb]{ .388,  .745,  .482}0.192 & \cellcolor[rgb]{ .388,  .745,  .482}1.785 & \cellcolor[rgb]{ .51,  .796,  .588}0.312 & \cellcolor[rgb]{ .388,  .745,  .482}2.347 \\
    Difficulty & \cellcolor[rgb]{ .988,  .988,  1}9.034 & \cellcolor[rgb]{ .78,  .906,  .824}0.155 & \cellcolor[rgb]{ .988,  .988,  1}0.185 & \cellcolor[rgb]{ .988,  .988,  1}1.680 & \cellcolor[rgb]{ .988,  .988,  1}0.302 & \cellcolor[rgb]{ .988,  .988,  1}2.271 \\
    Quality & \cellcolor[rgb]{ .396,  .749,  .486}9.289 & \cellcolor[rgb]{ .988,  .988,  1}0.154 & \cellcolor[rgb]{ .514,  .796,  .588}0.191 & \cellcolor[rgb]{ .537,  .808,  .612}1.759 & \cellcolor[rgb]{ .388,  .745,  .482}0.315 & \cellcolor[rgb]{ .435,  .765,  .525}2.341 \\
    \midrule
    \bottomrule
    \end{tabular}%
  \label{tab:data_filter}%
\end{table}%

\subsection{Data Filtering for Medical Reasoning}\label{sec:analysis_4.4}

Previous studies in general reasoning domains, such as mathematics, have demonstrated that filtering training data based on reasoning quality or problem difficulty can achieve comparable or superior performance relative to training on complete datasets, while significantly reducing computational cost~\cite{muennighoff2025s1, ye2025limo}.

We extend this investigation to the medical domain by applying two filtering regimes to the medical-o1 SFT corpus. As presented in Table~\ref{tab:data_filter}: \textbf{Quality-based filtering} retains only those examples whose CoT reasoning steps have been verified for resulting in the correct answer. This strategy preserves nearly all performance (Accuracy: 52.62\% vs.\ 52.79\%; $\Delta I$: 2.341 vs.\ 2.347) while discarding training tokens.
\textbf{Difficulty-based filtering} selects examples based on question difficulty. In contrast to the results in general reasoning, this approach degrades medical performance (Accuracy: 49.28\%; $\Delta I$: 2.271), indicating that factual correctness is paramount in knowledge-intensive settings.

In both the general and medical domains, quality‑based filtering delivers the best cost‑to‑performance ratio. But unlike mathematics, where difficulty‑based filtering can be neutral or beneficial, medical reasoning degrades when filtered purely by difficulty, underscoring the primacy of knowledge correctness in medical contexts. 

\newpage

\end{document}